\documentclass{article} 
\usepackage{iclr2027_conference,times}

\usepackage{amsmath,amsfonts,bm}

\def\eqref#1{equation~\ref{#1}}

\def\1{\bm{1}}

\DeclareMathAlphabet{\mathsfit}{\encodingdefault}{\sfdefault}{m}{sl}
\SetMathAlphabet{\mathsfit}{bold}{\encodingdefault}{\sfdefault}{bx}{n}

\usepackage{longtable}
\usepackage{booktabs}
\usepackage{array}
\usepackage{hyperref}
\usepackage{url}
\usepackage{amsmath}
\usepackage{amssymb}
\usepackage{amsthm}
\usepackage{placeins}
\usepackage{booktabs,graphicx,array,tabularx}
\usepackage{wrapfig}
\usepackage{capt-of}

\usepackage{xcolor}
\usepackage{colortbl}
\usepackage{nicematrix}

\usepackage{caption}
\newsavebox{\toyLeftBox}
\newsavebox{\toyRightBox}
\newsavebox{\toyLeftCaptionBox}
\newsavebox{\toyRightCaptionBox}
\newlength{\toyBodyPadding}
\newlength{\toyHeightDifference}
\newlength{\toyCaptionHeight}

\newcommand{\toyBodyRow}[1]{%
  \noalign{\vskip\toyBodyPadding}%
  #1\\%
  \noalign{\vskip\toyBodyPadding}%
}

\newcommand{\toyPreparePair}[4]{%
  \setlength{\toyBodyPadding}{0pt}%
  \sbox{\toyLeftBox}{%
    \begin{minipage}{.48\linewidth}\small#3\end{minipage}%
  }%
  \sbox{\toyRightBox}{%
    \begin{minipage}{.48\linewidth}\small#4\end{minipage}%
  }%
  \setlength{\toyHeightDifference}{\dimexpr
    \ht\toyLeftBox+\dp\toyLeftBox
    -\ht\toyRightBox-\dp\toyRightBox\relax}%
  \ifdim\toyHeightDifference>0pt
    \divide\toyHeightDifference by #2\relax
    \divide\toyHeightDifference by 2\relax
    \setlength{\toyBodyPadding}{\toyHeightDifference}%
    \sbox{\toyRightBox}{%
      \begin{minipage}{.48\linewidth}\small#4\end{minipage}%
    }%
  \else
    \ifdim\toyHeightDifference<0pt
      \setlength{\toyHeightDifference}{-\toyHeightDifference}%
      \divide\toyHeightDifference by #1\relax
      \divide\toyHeightDifference by 2\relax
      \setlength{\toyBodyPadding}{\toyHeightDifference}%
      \sbox{\toyLeftBox}{%
        \begin{minipage}{.48\linewidth}\small#3\end{minipage}%
      }%
    \fi
  \fi
}

\newcommand{\toyCaptionPair}[4]{%
  \sbox{\toyLeftCaptionBox}{%
    \begin{minipage}{.48\linewidth}
      \caption{#1}\label{#2}%
    \end{minipage}%
  }%
  \sbox{\toyRightCaptionBox}{%
    \begin{minipage}{.48\linewidth}
      \caption{#3}\label{#4}%
    \end{minipage}%
  }%
  \setlength{\toyCaptionHeight}{\dimexpr
    \ht\toyLeftCaptionBox+\dp\toyLeftCaptionBox\relax}%
  \ifdim\dimexpr
    \ht\toyRightCaptionBox+\dp\toyRightCaptionBox\relax
    >\toyCaptionHeight
    \setlength{\toyCaptionHeight}{\dimexpr
      \ht\toyRightCaptionBox+\dp\toyRightCaptionBox\relax}%
  \fi
  \begin{minipage}[t]{.48\linewidth}
    \vspace{0pt}
    \begin{minipage}[t][\toyCaptionHeight][t]{\linewidth}
      \vspace{0pt}\usebox{\toyLeftCaptionBox}
    \end{minipage}\par\nointerlineskip
    \usebox{\toyLeftBox}
  \end{minipage}\hfill%
  \begin{minipage}[t]{.48\linewidth}
    \vspace{0pt}
    \begin{minipage}[t][\toyCaptionHeight][t]{\linewidth}
      \vspace{0pt}\usebox{\toyRightCaptionBox}
    \end{minipage}\par\nointerlineskip
    \usebox{\toyRightBox}
  \end{minipage}%
}

\newcommand{\toyPanelPair}[2]{%
  \begin{minipage}[t]{.48\linewidth}
    \vspace{0pt}
    \centering
    \small
    \textbf{#1}\par\smallskip
    \nointerlineskip
    \usebox{\toyLeftBox}
  \end{minipage}\hfill%
  \begin{minipage}[t]{.48\linewidth}
    \vspace{0pt}
    \centering
    \small
    \textbf{#2}\par\smallskip
    \nointerlineskip
    \usebox{\toyRightBox}
  \end{minipage}%
}

\theoremstyle{definition}
\newtheorem*{crpodefinition}{Definition}

\theoremstyle{plain}
\newtheorem{rangeproposition}{Proposition}

\newcommand{\FinishToyWrap}{%
  \par
  \ifnum\value{WF@wrappedlines}>1\relax
    \begingroup
      \dimen0=\baselineskip
      \multiply\dimen0 by
        \numexpr\value{WF@wrappedlines}-1\relax
      \vskip\dimen0
    \endgroup
  \fi
  \WFclear
}

\usepackage{algorithm}
\usepackage{algpseudocode}

\title{Range-GRPO: Policy Optimization via Pairwise Relations among Reward Intervals}

\author{Ryunyi Lee, Kangjun Noh, Somin Kim, Heedong Kim, Kyungwoo Song$^{*}$ \\
Yonsei University
}

\iclrfinalcopy 
\begin{document}

\maketitle
\begingroup
\renewcommand{\thefootnote}{*}
\footnotetext{Corresponding author: kyungwoo.song@gmail.com}
\endgroup

\fancyhead[L]{Preprint}

\begin{abstract}
As the use of large language models (LLMs) expands, post-training has become
increasingly important for adapting them to downstream tasks.
However, obtaining reliable supervision remains costly,
especially in domains without reference answers or executable verifiers.
LLM-as-a-Judge provides scalable pseudo-rewards for unlabeled responses,
but a single point score does not explicitly represent reward uncertainty. This motivates representing pseudo-rewards as conformally calibrated
reward ranges. We propose \textbf{Range-GRPO}, a semi-supervised post-training framework
that combines limited labeled data with unlabeled prompts.
In Group Relative Policy Optimization (GRPO), learning signals depend on
relative reward comparisons within each rollout group.
The proposed objective compares reward ranges pairwise rather than
reducing them to point rewards, allowing interval uncertainty to affect
both the magnitude and direction of these signals.
Our theoretical analysis characterizes this distinction and shows that the proposed objective recovers the Dr.GRPO advantage when all reward ranges collapse
to points.
Empirically, Range-GRPO achieves the highest in-distribution and out-of-distribution average performance among the
evaluated semi-supervised methods while requiring fewer training resources.
\end{abstract}

\section{Introduction}

Adapting large language models (LLMs) to new tasks and domains requires
reliable supervision, yet obtaining it at scale remains difficult.
Existing benchmarks provide limited coverage of task distributions
\citep{kiela2021dynabench} and may suffer from data leakage
\citep{roberts2023cutoff,white2025livebench}.
These limitations motivate semi-supervised post-training methods
that use limited labeled data to learn from large collections of
unlabeled prompts. 
LLM-as-a-Judge (LaaJ) offers a scalable way to assign pseudo rewards to
unlabeled responses \citep{lee2023rlaif,bai2022constitutional}.
However, a judge may assign high scores to flawed reasoning and low
scores to valid but atypical responses.
It may also react to surface cues unrelated to correctness
\citep{ye2025justice,zheng2023judging,wu2025style}, but a single scalar score does not express the uncertainty of that evaluation.
Conformal calibration offers a promising approach to representing
this uncertainty through reward intervals \citep{sheng2025analyzing}.
In this setting, it measures the discrepancies between predicted and
reference rewards on a small labeled calibration set and uses the
empirical error distribution to construct intervals for new responses.
Under exchangeability, these intervals cover the reference reward
with marginal coverage of at least $1-\alpha$
\citep{angelopoulos2022gentleintroductionconformalprediction}.
This guarantee provides statistical support for the calibrated intervals,
but does not specify how to construct effective policy learning signals
from them.

We therefore seek an objective that uses calibrated intervals directly to determine relative credit among responses.
Some existing approaches predict task outputs as intervals and
derive rewards by comparing these intervals with known reference values \citep{lin2026bagen,liu2026rec,wu2026tempr1}. These comparisons cannot yield rewards for unlabeled responses
without the corresponding reference information. More closely related, Uncertainty-Aware Reward Modeling (UARM) follows a reweighting approach for Group
Relative Policy Optimization (GRPO) \citep{shao2024deepseekmath},
using the widths of calibrated intervals to scale advantages computed
from point rewards \citep{pan2026uncertainty}.
However, this rescaling can adjust advantage magnitudes but cannot
change the direction determined by the point rewards, even when those
rewards provide a misleading learning signal.

We address this gap with \textbf{Range-GRPO}, a semi-supervised
post-training framework that learns from unlabeled prompts using conformally calibrated reward intervals.
The proposed objective aggregates pairwise midpoint differences into
group-relative advantages, using the widths of both reward intervals
to weight the contribution of each comparison.
By changing the balance between positive and negative pairwise
contributions, this weighting can alter both advantage magnitudes
and signs even when the interval midpoints remain fixed.
When all reward intervals collapse to points, this objective recovers
the credit assignment of Dr.GRPO as a special case
\citep{liu2025understanding}.

As the policy learns, its response distribution can drift from the
initial labeled calibration distribution, motivating weighted
conformal inference to account for this shift.
To construct reward intervals under this drift, we adapt Robust
Weighted Conformal Inference (RoWCI), which performs shift-aware
LaaJ calibration in a low-cardinality judgment state space
\citep{noh2026reliable}.

Our main contributions are as follows:
\begin{itemize}
\item We propose \textbf{Range-GRPO}, to our knowledge the first
semi-supervised GRPO framework to construct on-policy learning
signals through pairwise comparisons of conformal reward intervals.

    \item We theoretically show that pairwise interval comparisons exploit range information beyond scalar reduction to determine both the magnitude and direction of learning signals, while retaining Dr.GRPO as the point-reward special case.

    \item Empirically, Range-GRPO outperforms the evaluated range-based methods on the synthetic task. On real reasoning benchmarks, it achieves the highest ID and OOD
average accuracies among the evaluated semi-supervised methods, with
relative improvements of \textbf{5.2\%} and \textbf{7.0\%} over the strongest baseline in our comparison,
respectively.
\end{itemize}

\section{Related work}

\subsection{Reward Supervision}
Reinforcement learning with verifiable rewards (RLVR) typically derives rewards from ground truth answers \citep{guo2025deepseek, ma2026general} or executable verifiers. Prior work also combines outcome rewards with signals derived
from internal model states \citep{sun2025detection}. Methods based on majority voting use agreement among unlabeled rollouts to construct pseudo labels \citep{wei2026first,zuo2026ttrl}. Another line of work uses LaaJ to assign rewards according to explicit rubrics
\citep{zhou2025breaking,gunjal2026rubrics}. TraPO, a semi-supervised RLVR framework, uses a small labeled set to select unlabeled samples whose learning trajectories align with those of labeled samples \citep{yang2026trapo}. Together, these approaches broaden the scope of available reward supervision beyond direct outcome verification. However, consensus rewards may reinforce incorrect answers, while point rewards from LLM judges may reflect biases unrelated
to response quality. Range-GRPO uses judge estimates with limited labeled calibration data
to construct conformal reward intervals that represent reward uncertainty on unlabeled rollouts.

\subsection{Conformal Calibration}

Conformal calibration is a promising approach to quantifying uncertainty
in LLM judge evaluations \citep{sheng2025analyzing}.
It constructs prediction intervals from labeled calibration data
and provides marginal coverage guarantees under exchangeability
\citep{angelopoulos2022gentleintroductionconformalprediction}.
However, policy updates can shift the response distribution away from
the calibration distribution, potentially violating the exchangeability assumption
underlying these guarantees.
Weighted Conformal Inference (WCI) addresses covariate shift by reweighting
calibration samples using density ratios \citep{tibshirani2019conformal}.
RoWCI, which we adopt, mitigates weight instability due to nuisance
variation in LLM judge inputs \citep{noh2026reliable}.
Recent studies have integrated conformal prediction into LLM training.
CONST selects prompts for annotation based on conformal prediction set
size \citep{xiao2026sample}. 
CFA uses conformal prediction sets to estimate
answer reliability and reweight preference pairs while retaining
the given preference labels \citep{chen2026conformal}. 
UARM uses calibrated reward interval widths to rescale GRPO advantages
derived from point rewards. However, its interval calibration procedure does not explicitly correct for shifts between the calibration and response distributions \citep{pan2026uncertainty}.
In contrast, Range-GRPO accounts for these shifts when constructing conformal reward intervals for unlabeled rollouts.

\subsection{Range-Based Learning Signals}
\label{sec:range_based_learning_signals}

Several studies derive scalar rewards by comparing predicted ranges
with task targets. BAGEN rewards narrow ranges that contain the realized cost of the
remaining trajectory \citep{lin2026bagen}.
REC-RL rewards predicted count ranges that contain the target, assigning larger rewards as both bounds approach it \citep{liu2026rec}.
TempR1 uses the overlap between predicted and ground-truth temporal
intervals as a reward signal \citep{wu2026tempr1}. Other studies use ranges to account for uncertainty in training
signals.
In its interval setting, Variance-Aware PPO samples scalar rewards
uniformly from a predicted reward range and uses the corresponding
variance to regularize policy updates \citep{banerjee2025towards}.
IRPM-Interval constructs Student's $t$ confidence intervals over
chosen and rejected score groups to define binary rollout rewards
\citep{song2026irpm}.
UARM uses the median quantile as a point reward
and the width of a conformally calibrated reward range to reweight response advantages without changing their signs \citep{pan2026uncertainty}.
Range-GRPO instead assigns relative credit within each rollout group
through pairwise comparisons of conformal reward intervals,
allowing reward uncertainty to influence both advantage magnitudes
and signs.

\section{Preliminaries}
\label{sec:preliminaries}

\subsection{Group Relative Policy Optimization}
\label{sec:grpo}

Group Relative Policy Optimization (GRPO) is a critic-free variant of
Proximal Policy Optimization (PPO) that estimates advantages by comparing the
rewards of multiple responses to the same prompt
\citep{shao2024deepseekmath}. For a prompt \(q\sim P_Q\), the old policy
samples \(G\) responses
\(o_i\sim\pi_{\theta_{\mathrm{old}}}(\cdot\mid q)\).
Under outcome supervision, GRPO computes
\(A_i=(r_i-\bar r)/\sigma_r\) from the response rewards
\(r_1,\ldots,r_G\), where
\(\bar r=G^{-1}\sum_j r_j\) and
\(\sigma_r=\operatorname{std}(r_1,\ldots,r_G)\). If $\sigma_r=0$, we set $A_i=0$.
For token \(k\), let
\(\rho_{i,k}=
\pi_\theta(o_{i,k}\mid q,o_{i,<k})/
\pi_{\theta_{\mathrm{old}}}(o_{i,k}\mid q,o_{i,<k})\).
GRPO applies the same response-level advantage \(A_i\) to
all tokens in \(o_i\) and maximizes
\begin{equation}
\label{eq:grpo_objective}
\begin{aligned}
\mathcal J_{\mathrm{GRPO}}(\theta)
=
\mathbb E\Bigg[
\frac{1}{G}
\sum_{i=1}^{G}
\frac{1}{|o_i|}
\sum_{k=1}^{|o_i|}
\Big\{
\min\!\big(
\rho_{i,k}A_i,\,
\operatorname{clip}(\rho_{i,k},1-\epsilon,1+\epsilon)A_i
\big)
-
\beta D_{i,k}^{\mathrm{KL}}
\Big\}
\Bigg].
\end{aligned}
\end{equation}
In Eq.~(\ref{eq:grpo_objective}), \(D_{i,k}^{\mathrm{KL}}\) penalizes
deviation from a fixed reference policy \(\pi_{\mathrm{ref}}\), while
\(\epsilon\) and \(\beta\) control the clipping range and KL
regularization strength, respectively.
Since \(A_i\) determines the direction and magnitude of the learning
signal for \(o_i\), the quality of advantage estimates directly affects
policy updates. We build on Dr.GRPO, a variant of GRPO that computes
\(A_i\) without normalization by \(\sigma_r\) and removes
response length normalization from the objective
\citep{liu2025understanding}. For convenience, Appendix~\ref{sec:notation} summarizes the notation
used throughout the paper.

\subsection{Reward Range Construction}
\label{sec:conformal_reward_range}

We adapt Robust Weighted Conformal Inference (\textsc{RoWCI}) to construct
reward ranges for responses generated by the current policy when reference rewards are unavailable \citep{noh2026reliable}. At a fixed optimization step \(t\),
let \(X=(q,o)\), where \(o\sim\pi_{\theta_t}(\cdot\mid q)\), and let
\(Y\in[0,1]\) denote the corresponding reference reward. Let \(P_{\mathrm{src}}\) and \(P_t\)
denote the source and target distributions over \((X,Y)\), respectively.
We observe both \(X\) and \(Y\) in the source data but only \(X\) in rollouts
from \(\pi_{\theta_t}\). Following RoWCI, we represent $X$ using a low-cardinality judgment state
$Z=\phi(X)$ to reduce sensitivity to surface cues in LaaJ evaluations. When some judgment-state components depend on relationships with other
responses in the same rollout group, the state extractor additionally
conditions on the corresponding group.
We suppress this group-context argument in the notation $\phi(X)$.
We perform distribution-shift reweighting in the judgment-state space
and assume
\(P_{\mathrm{src}}(Y\mid Z)\approx P_t(Y\mid Z)\) and
\(P_t^Z\ll P_{\mathrm{src}}^Z\), where \(P_{\mathrm{src}}^Z\) and \(P_t^Z\)
denote the source and target distributions of \(Z\), respectively.

Given a fixed state scorer \(s_Z:\mathcal Z\to[0,1]\), we compute the
nonconformity score \(E_j^Z=|s_Z(Z_j)-Y_j|\) for each source calibration pair
\((X_j,Y_j)\), where \(Z_j=\phi(X_j)\). A binary domain classifier outputs
\(\widehat p_t^{\mathrm{tgt}}(z)\approx\Pr(C=1\mid Z=z)\), where \(C=0\) and \(C=1\)
denote the source and target domains, respectively. We compute the calibration weights as
\(\widehat w_t(z)
=
\frac{m_{\mathrm{src}}}{m_t}
\frac{\widehat p_t^{\mathrm{tgt}}(z)}{1-\widehat p_t^{\mathrm{tgt}}(z)}\),
where \(m_{\mathrm{src}}\) and \(m_t\) are the numbers of source and target samples
used to train the domain classifier. The reward range is constructed as
$\widehat{\mathcal I}_{\alpha,t}(X)
=
\big[
s_Z(\phi(X))-\widehat\tau_{\alpha,t}(\phi(X)),\,
s_Z(\phi(X))+\widehat\tau_{\alpha,t}(\phi(X))
\big]\cap[0,1]$,
where $\widehat\tau_{\alpha,t}(z)$ is the weighted conformal cutoff
for nominal marginal coverage $1-\alpha$.
For the weighted cutoff, we include a point mass at \(+\infty\)
with weight \(\widehat w_t(z)\) for the new response.
In the main experiments, we instead normalize the nonconformity scores
to account for variation in reward uncertainty across scorer outputs,
as described in Section~\ref{sec:range-grpo-implementation}.
Range-GRPO constructs a reward interval for each of the $G$ responses
to a prompt.

\begin{figure}[t]
    \centering
    \includegraphics[
        width=\linewidth,
        trim={0 7pt 0 0},
        clip
    ]{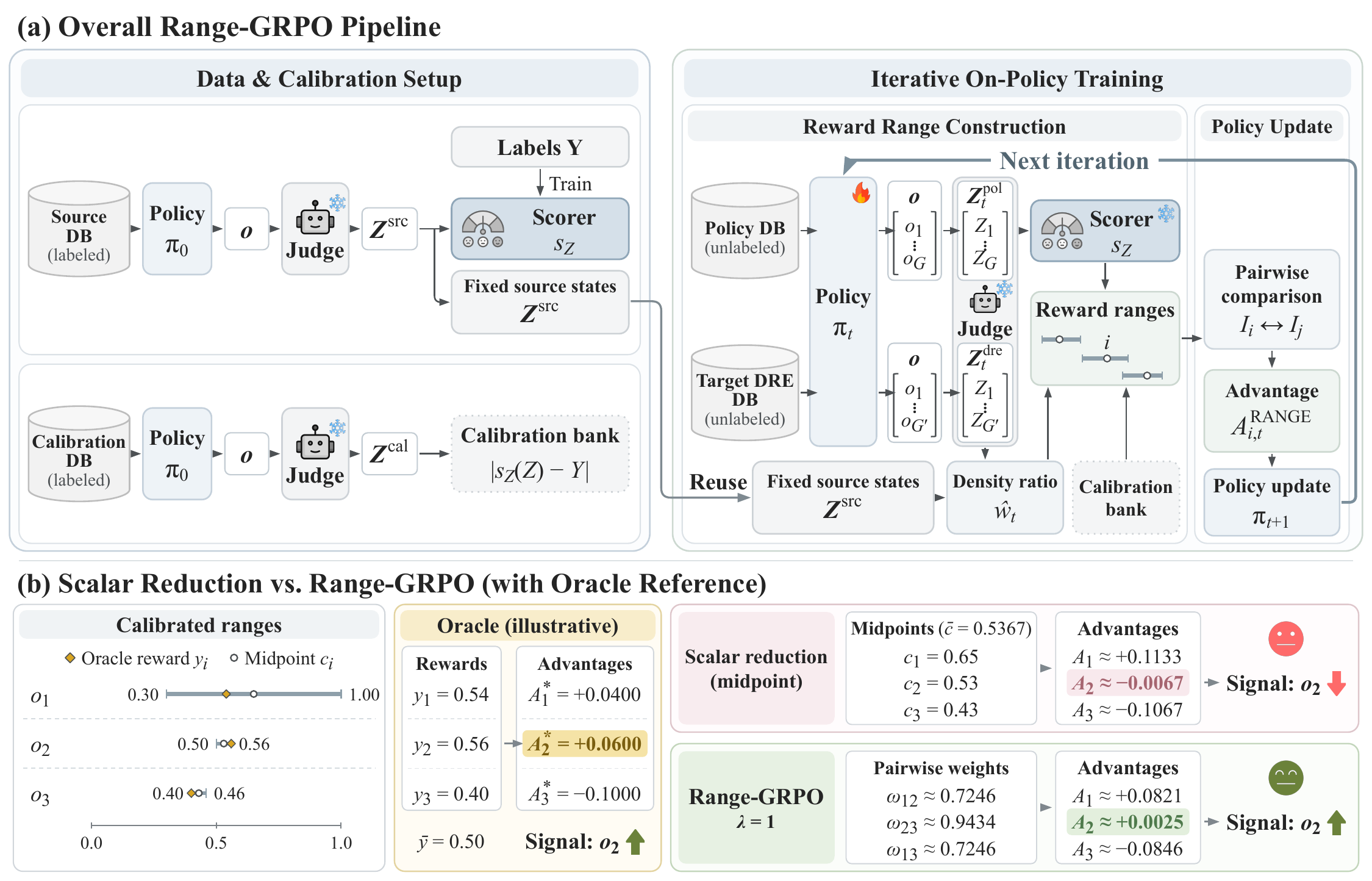}
    \caption{
        \textbf{Overview of Range-GRPO and its learning signals.}
        \textbf{(a)} At initialization, Range-GRPO fits a state scorer on
        labeled source rollouts and computes nonconformity scores on
        calibration rollouts.
        Each iteration generates policy and target DRE rollouts with
        $G$ and $G'$ responses, respectively, and updates density ratios using
        the target states and reused source states.
        The density ratios, fixed scorer, and calibration scores are
        used to construct reward ranges, from which pairwise interval comparisons
        construct the advantages for policy updates.
        \textbf{(b)} An illustrative comparison uses identical reward intervals
        for both methods.
        Although the interval for $o_2$ lies above that for $o_3$ with the
        same width, midpoint reduction assigns it a negative advantage under Dr.GRPO.
        Range-GRPO instead downweights pairwise contributions involving wider,
        more uncertain intervals, turning this advantage positive and aligning the
        direction of the learning signal with the oracle preference.
    }
    \label{fig:range-grpo-pipeline}
\end{figure}

\section{Method}
\label{sec:method}

We formulate the Range-GRPO objective and characterize its pairwise
credit assignment relative to reducing each reward range to a point
reward at a fixed relative position.
Range-GRPO reduces to Dr.GRPO when all reward ranges collapse to points.
Finally, we describe the on-policy training procedure that integrates
this objective with shift-aware conformal reward calibration, as illustrated
in Figure~\ref{fig:range-grpo-pipeline}.

\subsection{Range-GRPO Objective}
\label{sec:range_grpo_objective}

At iteration \(t\), the current policy
\(\pi_t:=\pi_{\theta_t}\) generates \(G\) responses
\(o_i\sim\pi_t(\cdot\mid q)\) for each prompt \(q\).
Each response is associated with a calibrated reward interval
\(\mathcal I_i:=\widehat{\mathcal I}_{\alpha,t}(X_i)=[L_i,U_i]\),
whose center and half-width are
\(c_i=(L_i+U_i)/2\) and \(h_i=(U_i-L_i)/2\), respectively.
Below, we suppress the dependence on \(q\), \(t\), and \(\alpha\),
with \(i\) indexing the response of interest and \(j\) a comparison
response.

\begin{crpodefinition}[Range-GRPO objective]
\label{def:crpo_objective}
For each pair of responses $i\neq j$, define the comparison weight
$\omega_{ij}=\frac{\lambda}{h_i+h_j+\lambda}$, where $\lambda>0$.
The corresponding pairwise signal is
$g_{ij}=\omega_{ij}(c_i-c_j)$.
Range-GRPO aggregates these signals into the response-level advantage
\begin{equation}
    A_i^{\mathrm{RANGE}}
    =
    \frac{1}{G}\sum_{j\neq i}g_{ij}
    =
    \frac{1}{G}\sum_{j\neq i}\omega_{ij}(c_i-c_j).
    \label{eq:range_final_advantage}
\end{equation}
We define the Range-GRPO objective by replacing the Dr.GRPO
response-level advantage \(A_i^{\mathrm{Dr.GRPO}}\) with
\(A_i^{\mathrm{RANGE}}\), while retaining the remaining
objective structure.
\end{crpodefinition}

The center difference $c_i-c_j$ determines the direction of the
pairwise signal, while $\omega_{ij}$ scales its magnitude.
We use reward interval width as a proxy for uncertainty in the
reward estimate.
Importantly, the reliability of a pairwise comparison depends on
the uncertainty of both responses.
A wide interval for either response reduces confidence in the
difference between their centers, even when the other interval
is narrow. We therefore define $\omega_{ij}$ using the combined half-width
$h_i+h_j$, so that uncertainty in either response attenuates
the corresponding pairwise contribution.
The reference width $\lambda$ controls the sensitivity of this
attenuation to interval width.

\begin{rangeproposition}[Range-GRPO under Zero Interval Widths]
\label{prop:crpo_properties}
Suppose $\mathcal{I}_i=[r_i,r_i]$ for all $i=1,\ldots,G$.
Then $\omega_{ij}=1$ for every pair, and
\begin{equation}
    A_i^{\mathrm{RANGE}}
    =
    \frac{1}{G}\sum_{j\neq i}(r_i-r_j)
    =
    r_i-\bar r
    =
    A_i^{\mathrm{Dr.GRPO}}.
    \label{eq:crpo_exact_zero_width}
\end{equation}
\end{rangeproposition}

Thus, Range-GRPO reduces exactly to Dr.GRPO when all reward intervals
collapse to points.

\subsection{Preserving Range Information through Pairwise Credit Assignment}
\label{sec:pairwise_range_credit}

A point reduction maps each reward interval
$\mathcal{I}_i=[L_i,U_i]$ to a single scalar reward.
For a fixed relative position $\xi\in[0,1]$, define
$r_i^{\mathrm{pt}}=\xi L_i+(1-\xi)U_i=c_i+(1-2\xi)h_i$.
The lower bound (LB), midpoint, and upper bound (UB) correspond to
$\xi=1$, $\xi=1/2$, and $\xi=0$, respectively.
Since different combinations of $c_i$ and $h_i$ can yield the same
$r_i^{\mathrm{pt}}$, the scalar reward alone does not uniquely determine the underlying interval.
This limitation becomes explicit under midpoint reduction, where
$r_i^{\mathrm{pt}}=c_i$ is independent of $h_i$.

We examine how this information loss affects group-relative credit assignment.
For a rollout group $\mathcal G$, let
$A_i^{\mathrm{pt}}(\mathcal G)
=
r_i^{\mathrm{pt}}-\bar r^{\mathrm{pt}}(\mathcal G)$,
where $\bar r^{\mathrm{pt}}(\mathcal G)$ denotes the mean point reward
in $\mathcal G$.
In the following, we isolate the effect of interval width by holding
midpoint rewards fixed.

\begin{rangeproposition}[Effect of Pairwise Range Information on Credit Assignment]
\label{prop:range_width_direction}

Let the point rewards be interval midpoints $(\xi=1/2)$ and consider
two rollout groups with $G=3$,
$\mathcal G_j=\{i,\ell,j\}$ and
$\mathcal G_k=\{i,\ell,k\}$, where
$r_j^{\mathrm{pt}}=r_k^{\mathrm{pt}}>
r_i^{\mathrm{pt}}>r_\ell^{\mathrm{pt}}$ and $h_j<h_k$. The intervals of $i$ and $\ell$ remain fixed across the two groups.
Suppose $|g_{ik}|<g_{i\ell}<|g_{ij}|$. Then
\[
\begin{aligned}
\textup{(1)}\quad
&A_i^{\mathrm{RANGE}}(\mathcal G_j)
<0<
A_i^{\mathrm{RANGE}}(\mathcal G_k),\\
\textup{(2)}\quad
&A_i^{\mathrm{pt}}(\mathcal G_j)
=
A_i^{\mathrm{pt}}(\mathcal G_k),\\
\textup{(3)}\quad
&\operatorname{sign}\!\left(A_i^{\mathrm{wt}}(\mathcal G_j)\right)
=
\operatorname{sign}\!\left(A_i^{\mathrm{wt}}(\mathcal G_k)\right),
\end{aligned}
\]
where
$A_i^{\mathrm{wt}}(\mathcal G)
=\eta_i(\mathcal G)A_i^{\mathrm{pt}}(\mathcal G)$
with $\eta_i(\mathcal G)>0$.
\end{rangeproposition}

Proposition~\ref{prop:range_width_direction} shows that the advantage
sign for response $i$ under Range-GRPO can reverse as interval
widths change, even when the midpoint rewards remain fixed. For response $i$, $c_j=c_k>c_i>c_\ell$ implies $g_{i\ell}>0$ and $g_{ij},g_{ik}<0$.
Since $h_j<h_k$, the wider interval of $k$ yields
$\omega_{ik}<\omega_{ij}$, thereby reducing the magnitude of
its negative pairwise contribution.

The resulting advantages of response $i$ are
$A_i^{\mathrm{RANGE}}(\mathcal G_j)
=\frac{1}{G}(g_{i\ell}+g_{ij})$
and
$A_i^{\mathrm{RANGE}}(\mathcal G_k)
=\frac{1}{G}(g_{i\ell}+g_{ik})$.
The assumed inequality gives
$g_{i\ell}+g_{ij}<0<g_{i\ell}+g_{ik}$, establishing
result~\textup{(1)}.
Under midpoint reduction, $j$ and $k$ retain the same point reward
despite having different interval widths, yielding the same
point-reduction advantage in result~\textup{(2)}.
Finally, $\eta_i(\mathcal G_j)$ and $\eta_i(\mathcal G_k)$ are both
positive.
Response-level reweighting therefore preserves the sign of the common
point-reduction advantage, so result~\textup{(3)} follows.

UARM~\citep{pan2026uncertainty} provides a concrete example of the
response-level weighting in statement~\textup{(3)}.
For a matched comparison with Range-GRPO, we use the same reward intervals and set
$r_i=c_i$ for every response.
The UARM reweighting rule then gives
\begin{equation}
    A_i^{\mathrm{UARM}}(\mathcal G)
    =
    \eta_i^{\mathrm{UARM}}
    A_i^{\mathrm{pt}}(\mathcal G),
    \qquad
    \eta_i^{\mathrm{UARM}}
    =
    \frac{\sigma_{\mathrm{signal}}}
         {\sigma_{\mathrm{signal}}^2+
          \sigma_{\mathrm{noise},i}^2}
    >0.
    \label{eq:uarm_matched_weighting}
\end{equation}
The scaling factor can be written as
$\frac{\sigma_{\mathrm{signal}}^2}
{\sigma_{\mathrm{signal}}^2+\sigma_{\mathrm{noise},i}^2}
\frac{1}{\sigma_{\mathrm{signal}}}
=
\eta_i^{\mathrm{UARM}}>0$,
since $\sigma_{\mathrm{signal}}>0$.
The positive scaling factor can change the advantage magnitude,
but the centered midpoint reward still determines its sign.
Under this matched comparison, UARM satisfies
statement~\textup{(3)}.
In contrast, Range-GRPO applies interval-dependent weights to the
individual pairwise signals before aggregation, which allows the
advantage sign to change as shown in
Proposition~\ref{prop:range_width_direction}.

\begin{algorithm}[t]
\caption{The Range-GRPO Algorithm}
\label{alg:range-grpo}
\small
\begin{algorithmic}[1]

\State \textbf{Input:}
Prompt sets
$\mathcal D_U^{\mathrm{pol}},
\mathcal D_L^{\mathrm{src}},
\mathcal D_L^{\mathrm{cal}},
\mathcal D_U^{\mathrm{dre}}$;
initial policy $\pi_0$; frozen judge $J$

\State \textbf{Hyperparameters:}
miscoverage rate $\alpha$; reference width $\lambda$;
group size $G$; number of training iterations $T$

\Statex \textit{Stage 1: Data and Calibration Setup}

\State Generate source rollouts from $\mathcal D_L^{\mathrm{src}}$
and calibration rollouts from $\mathcal D_L^{\mathrm{cal}}$
using $\pi_0$

\State Obtain reference rewards
$\mathcal Y^{\mathrm{src}}$ and $\mathcal Y^{\mathrm{cal}}$
from the labeled data

\State Extract source states $\mathcal Z^{\mathrm{src}}$
and calibration states
$\mathcal Z^{\mathrm{cal}}=\{Z_j^{\mathrm{cal}}\}_{j=1}^{n_{\mathrm{cal}}}$
with $J$

\State Fit the state scorer
$s_Z:\mathcal Z\rightarrow[0,1]$ on
$(\mathcal Z^{\mathrm{src}},\mathcal Y^{\mathrm{src}})$

\State Compute calibration nonconformity scores
$\{E_j^Z\}_{j=1}^{n_{\mathrm{cal}}}$
as defined in Section~\ref{sec:conformal_reward_range}

\State Fix $s_Z$ and retain $\mathcal Z^{\mathrm{src}}$ for DRE and
$\{(Z_j^{\mathrm{cal}},E_j^Z)\}_{j=1}^{n_{\mathrm{cal}}}$
for calibration throughout training

\For{$t=0,\ldots,T-1$}

    \Statex \textit{Stage 2: On-Policy Reward Range Construction}

    \State Sample a policy prompt batch $\mathcal Q_t$
    from $\mathcal D_U^{\mathrm{pol}}$

    \State Generate $G$ rollouts per prompt in $\mathcal Q_t$
    using $\pi_t$

    \State Sample fresh target DRE rollouts from $\pi_t$
    on $\mathcal D_U^{\mathrm{dre}}$

    \State Extract
    $\mathcal Z_t^{\mathrm{pol}}$ and $\mathcal Z_t^{\mathrm{dre}}$
    with $J$

    \State Fit a binary domain classifier on
    $\mathcal Z^{\mathrm{src}}$ and $\mathcal Z_t^{\mathrm{dre}}$
    to estimate the target-to-source density ratio
    $\widehat w_t(\cdot)$
    
    \For{each policy rollout $X_i$ with state $Z_{i,t}^{\mathrm{pol}}$}
    
        \State Compute the weighted conformal cutoff
        $\widehat\tau_{\alpha,t}(Z_{i,t}^{\mathrm{pol}})$ as defined in
        Section~\ref{sec:conformal_reward_range}
    
        \State $\widehat{\mathcal I}_{\alpha,t}(X_i)
        \gets
        \big[
        s_Z(Z_{i,t}^{\mathrm{pol}})
        -\widehat\tau_{\alpha,t}(Z_{i,t}^{\mathrm{pol}}),
        s_Z(Z_{i,t}^{\mathrm{pol}})
        +\widehat\tau_{\alpha,t}(Z_{i,t}^{\mathrm{pol}})
        \big]\cap[0,1]$

    \EndFor

    \Statex \textit{Stage 3: Range-GRPO Policy Update}

    \State Compute the advantage $A_{i,t}^{\mathrm{RANGE}}$
    for each rollout within each prompt group using
    Eq.~(\ref{eq:range_final_advantage})

    \State Update $\pi_t$ by optimizing the Range-GRPO objective
    on the policy rollouts to obtain $\pi_{t+1}$
    
\EndFor

\State \textbf{Output:} final policy $\pi_T$

\end{algorithmic}
\end{algorithm}

\subsection{On-Policy Training Procedure}
\label{sec:range-grpo-method}

Range-GRPO combines the proposed objective with the reward range
construction in Section~\ref{sec:conformal_reward_range}
for on-policy training.
At initialization, it fits the state scorer and computes
nonconformity scores using rollouts from the initial policy.
Each iteration then generates fresh policy and target DRE rollouts,
updates the density ratio weights, and constructs reward ranges for
the policy update.

\noindent\textbf{Calibration Setup.}
We use four disjoint sets of prompts: an unlabeled policy set
$\mathcal D_U^{\mathrm{pol}}$, a labeled source set
$\mathcal D_L^{\mathrm{src}}$, a labeled calibration set
$\mathcal D_L^{\mathrm{cal}}$, and an unlabeled target set
$\mathcal D_U^{\mathrm{dre}}$ for density ratio estimation (DRE).
The source and calibration sets come from the same labeled prompt
distribution, whereas the policy and target DRE sets come from the
same unlabeled prompt distribution.

The initial policy $\pi_0$ generates rollouts from
$\mathcal D_L^{\mathrm{src}}$ and $\mathcal D_L^{\mathrm{cal}}$.
Let $\phi$ denote the fixed state-extraction procedure implemented
by the frozen judge $J$, so that each rollout $X_i$ is mapped to
a judgment state $Z_i=\phi(X_i)$.
The same state-extraction procedure is applied to all four sets
throughout training. Each source and calibration rollout receives a reference reward
$Y_i\in[0,1]$ from the labeled data.
We fit the state scorer $s_Z$ using the source states and rewards
and compute calibration nonconformity scores as described in
Section~\ref{sec:conformal_reward_range}.
The state scorer, source and calibration states, and calibration nonconformity scores remain fixed throughout training.

\noindent\textbf{On-Policy Training.}
At iteration $t$, the current policy $\pi_t$ and the unlabeled
prompt distribution induce the target distribution $P_t$.
We sample a batch of prompts from $\mathcal D_U^{\mathrm{pol}}$
and generate $G$ rollouts per prompt.
Using the same policy and sampling settings, we also generate
$G'$ rollouts per target DRE prompt from the fixed set
$\mathcal D_U^{\mathrm{dre}}$.
The judge $J$ extracts states from both sets of rollouts,
but we exclude the target DRE rollouts from policy updates.

We fit a binary domain classifier on the fixed source states and
the current target DRE states, without using correctness rewards,
to update the estimated state density ratios $\widehat w_t$. Using these weights with the fixed calibration states
and nonconformity scores, we construct a reward range
$\widehat{\mathcal I}_{\alpha,t}(X_i)$ for each policy rollout $X_i$.
For each prompt, we compute advantages by comparing its $G$
reward ranges pairwise using Eq.~(\ref{eq:range_final_advantage}).
We then optimize the Range-GRPO objective on the policy rollouts
using these advantages to obtain $\pi_{t+1}$. Algorithm~\ref{alg:range-grpo} summarizes the procedure.

\section{Experiments}

\subsection{Toy Experiment}
\label{sec:crpotoy-main}

\textbf{Setup.}
We adapt a synthetic task from prior work \citep{lee2026semi}
into a candidate selection task.
GPT-2-small \citep{radford2019language} serves as the policy,
and the prompt does not include the ranking rule.
The rank of the selected candidate determines the true reward.
Accuracy measures how often the policy selects the candidate
with rank~1, which receives the highest reward.
The labeled training pool serves only to fit the state scorer and calibrate reward intervals.

\begin{table*}[t]
\centering
\caption{
\textbf{Range-GRPO achieves the highest mean test accuracy among
the compared range-based methods.}
We select the checkpoint with the highest
validation accuracy over ten training epochs and report its
test accuracy.
All values are means $\pm$ sample standard deviations across
three seeds.
Cov. and Width summarize empirical coverage and mean interval width
on unlabeled responses over the full training phase.
We set the miscoverage rate to $\alpha=0.1$.
Bold and underlined values indicate the best and second best results,
respectively, for accuracy and interval width.
}
\label{tab:crpotoy-main}

\fontsize{8.5}{10.2}\selectfont\normalfont
\setlength{\tabcolsep}{1.0pt}
\renewcommand{\arraystretch}{1.28}

\setlength{\aboverulesep}{0pt}
\setlength{\belowrulesep}{0pt}
\setlength{\belowbottomsep}{0pt}

\begin{NiceTabular*}{\linewidth}{
    @{\extracolsep{\fill}}
    l
    @{\hspace{10pt}}
    l
    c
    @{\hspace{10pt}}
    c
    c
    @{}
}
\CodeBefore
    \rectanglecolor{gray!10}{2-1}{2-5}
    \rectanglecolor{gray!10}{9-1}{9-5}
\Body

\toprule
\textbf{Advantage}
& \textbf{Method}
& \textbf{Acc. (\%)} $\uparrow$
& \textbf{Cov. (\%)}
& \textbf{Width} $\downarrow$ \\

\midrule
\multicolumn{5}{c}{\textbf{Scalar reductions}} \\
\midrule

\Block[l]{3-1}{GRPO}
& LB
& $27.63 \pm 2.84$
& $93.69 \pm 0.65$
& $0.3942 \pm 0.0098$ \\

& Midpoint
& $23.97 \pm 7.61$
& $93.95 \pm 1.47$
& $0.4070 \pm 0.0120$ \\

& UB
& $24.60 \pm 2.71$
& $92.64 \pm 0.89$
& $0.3930 \pm 0.0122$ \\

\midrule

\Block[l]{3-1}{Dr.GRPO}
& LB
& $30.33 \pm 7.68$
& $93.29 \pm 0.27$
& $0.3753 \pm 0.0126$ \\

& Midpoint
& $\underline{31.77 \pm 14.11}$
& $93.02 \pm 0.11$
& $\underline{0.3699 \pm 0.0094}$ \\

& UB
& $29.83 \pm 4.84$
& $93.35 \pm 0.75$
& $0.3785 \pm 0.0136$ \\

\midrule
\multicolumn{5}{c}{\textbf{Main comparison}} \\
\midrule

\multicolumn{2}{c}{UARM}
& $20.43 \pm 3.61$
& $94.41 \pm 0.23$
& $0.4311 \pm 0.0090$ \\

\multicolumn{2}{c}{\textbf{Range-GRPO (Ours)}}
& $\boldsymbol{39.40 \pm 7.70}$
& $93.05 \pm 0.70$
& $\boldsymbol{0.3595 \pm 0.0206}$ \\

\bottomrule
\end{NiceTabular*}

\end{table*}

We compare Range-GRPO with LB, Midpoint, UB, and UARM.
LB, Midpoint, and UB have separate variants using
the GRPO \citep{shao2024deepseekmath} and
Dr.GRPO \citep{liu2025understanding} advantage formulas.
All methods share the data allocation and use the same on-policy
reward range construction described in
Section~\ref{sec:range-grpo-method}.
This controlled comparison focuses on how the methods derive learning
signals from reward ranges.
We use the true rewards of unlabeled responses only to compute
empirical coverage.

All methods draw from the same labeled and unlabeled pools and
have equal budgets for total rollout generation, including
auxiliary generation.
We train all methods for ten epochs with $G=8$.

\textbf{Results.}
Table~\ref{tab:crpotoy-main} shows that Range-GRPO achieves
the highest mean test accuracy ($39.40\%$) among the compared
methods, exceeding the best scalar reduction result by
$7.63$ percentage points.
All methods achieve mean empirical coverage above the nominal
$90\%$ level on unlabeled responses over the full training phase.
Range-GRPO also yields the smallest mean interval width
($0.3595$).
Appendix~\ref{sec:crpo-no-warmup-appendix} provides additional
training details and validation curves.

\subsection{Main Experiments}
\label{sec:main_experiments}

\begin{table}[t]
\centering
\caption{\textbf{Range-GRPO achieves the highest ID and OOD  averages among semi-supervised methods.} We report published baseline results from TraPO
\citep{yang2026trapo}, while methods marked with $^{*}$ are evaluated under our experimental setting. For the reproduced TraPO configurations, values in parentheses indicate
the entropy regularization coefficients. All training methods use 1,024 (1K) ID and 1,024 (1K) OOD prompts. Semi-supervised methods use labeled ID prompts and unlabeled OOD prompts, whereas the fully supervised model uses labels for all 2K prompts. GPQA$^{\dagger}$ denotes GPQA-diamond. Bold and underlined scores indicate the best and second-best results among semi-supervised methods, respectively, with ties receiving the same formatting. The fully supervised result is provided for reference only and excluded from this ranking. 
}
\label{tab:ood_unlabeled}

\fontsize{8.5}{11}\selectfont\normalfont
\setlength{\tabcolsep}{1.0pt}
\renewcommand{\arraystretch}{1.30}

\setlength{\aboverulesep}{0pt}
\setlength{\belowrulesep}{0pt}
\setlength{\belowbottomsep}{0pt}

\resizebox{\linewidth}{!}{%
\begin{NiceTabular}{
    @{}
    l
    c
    c
    c
    c
    c
    @{\hspace{4pt}}
    c
    @{\hspace{7pt}}
    c
    c
    c
    @{\hspace{4pt}}
    c
    @{}
}
\CodeBefore
    \rectanglecolor{gray!10}{3-1}{3-11}
    \rectanglecolor{gray!10}{6-1}{6-11}
    \rectanglecolor{gray!10}{8-1}{8-11}
\Body

\toprule
& \multicolumn{6}{c}{\textbf{In-Distribution}}
& \multicolumn{4}{c}{\textbf{Out-of-Distribution}} \\
\cmidrule(lr){2-7}
\cmidrule(lr){8-11}

Model / Method
& AIME 24/25
& AMC
& MATH-500
& Minerva
& Olympiad
& \textbf{Avg.}
& ARC-c
& GPQA$^{\dagger}$
& MMLU-Pro
& \textbf{Avg.} \\

\midrule
\multicolumn{11}{c}{\textbf{Backbone}} \\
\midrule

Qwen-Base
& 11.5/4.9
& 31.3
& 43.6
& 7.4
& 15.6
& 19.0
& 18.2
& 11.1
& 16.9
& 15.4 \\

Qwen-Instruct
& 12.5/10.2
& 48.5
& 80.4
& 32.7
& 41.0
& 37.6
& 70.3
& 24.7
& 34.1
& 43.0 \\

\midrule
\multicolumn{11}{c}{\textbf{Fully supervised}} \\
\midrule

\textcolor{gray!70}{Supervised (2K)}
& \textcolor{gray!70}{17.3/12.4}
& \textcolor{gray!70}{56.8}
& \textcolor{gray!70}{81.4}
& \textcolor{gray!70}{38.6}
& \textcolor{gray!70}{44.8}
& \textcolor{gray!70}{41.9}
& \textcolor{gray!70}{82.0}
& \textcolor{gray!70}{38.9}
& \textcolor{gray!70}{52.4}
& \textcolor{gray!70}{57.8} \\

\midrule
\multicolumn{11}{c}{\textbf{Semi-supervised}} \\
\midrule

TTRL
& 16.4/\underline{13.6}
& 49.9
& 66.9
& 26.5
& 37.8
& 35.2
& 62.0
& 31.8
& 43.5
& 45.8 \\

Self-certainty
& 16.0/10.9
& 53.0
& 78.4
& 34.2
& 39.0
& 38.6
& 77.1
& 32.8
& \underline{45.7}
& 51.9 \\

Token-level Entropy
& \underline{17.7}/11.0
& 51.7
& 77.0
& 33.1
& 41.0
& 38.6
& 76.5
& 30.8
& 44.7
& 50.7 \\

Sentence-level Entropy
& 15.7/10.0
& 51.4
& 77.4
& \underline{34.9}
& 37.5
& 37.8
& 75.1
& 31.3
& 44.3
& 50.2 \\

TraPO (entropy=0.001)$^{*}$
& \underline{17.7}/12.2
& \underline{55.9}
& 79.2
& \textbf{37.5}
& \underline{41.3}
& \underline{40.6}
& \underline{80.4}
& 34.8
& 43.5
& \underline{52.9} \\

UARM$^{*}$
& 17.1/9.5
& 53.2
& \underline{80.4}
& 33.5
& 40.9
& 39.1
& 78.9
& \underline{36.9}
& 34.2
& 50.0 \\

\specialrule{0.4pt}{2pt}{1.5pt}
\specialrule{0.4pt}{0pt}{2pt}

\textbf{Range-GRPO (ours), $\lambda=1$}
& \textbf{20.6}/\textbf{14.4}
& \textbf{57.0}
& \textbf{83.0}
& 33.8
& \textbf{47.3}
& \textbf{42.7}
& \textbf{80.9}
& \textbf{40.4}
& \textbf{48.4}
& \textbf{56.6} \\

\bottomrule
\end{NiceTabular}%
}

\end{table}

We evaluate Range-GRPO on established reasoning benchmarks
to complement the synthetic toy experiment. These experiments examine whether Range-GRPO can effectively learn
from unlabeled prompts with limited labeled supervision and generalize
beyond the labeled source domain. Appendix~\ref{sec:experimental_details} provides implementation details. 

\noindent\textbf{Baselines.}
We adapt the training and evaluation setting used for
Table~2 in \citet{yang2026trapo}, without increasing the
data or policy rollout budgets.
The comparison includes the semi-supervised results reported
in that table for TTRL \citep{zuo2026ttrl},
Self-certainty \citep{zhao2026learning}, and the Token-level
Entropy and Sentence-level Entropy baselines
\citep{agarwal2026unreasonable}.
We additionally reproduce TraPO under our experimental setting
and implement UARM \citep{pan2026uncertainty} with a quantile
reward model as a direct baseline.

\noindent\textbf{Training Setup.}
For policy training, we use Qwen2.5-Math-7B
\citep{yang2024qwen2} and generate $G=8$ responses per prompt.
The training data comprise 1,024 labeled mathematics prompts
from OpenR1-Math-220k \citep{face2025open} and 1,024 unlabeled
MMLU-Pro prompts \citep{wang2024mmlu}. For Range-GRPO and UARM, we reserve part of the training data
for their auxiliary components, including reward estimation,
calibration, and density ratio estimation (DRE) where applicable.
Training follows the TraPO schedule of ten labeled warmup epochs
followed by five epochs using both labeled and unlabeled prompts.
Labeled updates use binary correctness rewards based on reference
answers, while each method defines its own learning signals
for unlabeled updates.

\noindent\textbf{Reward Range Construction.}
After labeled warmup and before unlabeled policy updates begin,
we use the resulting policy to generate the source and calibration
responses for reward range construction. A frozen Qwen3-4B-Instruct \citep{yang2025qwen3} judge extracts judgment states for these responses. We then fit a lightweight XGBoost scorer on the source states and compute calibration nonconformity scores. The scorer and calibration scores remain fixed during subsequent
training, while density ratios are updated at each iteration using
fresh target responses from the current policy.
For each unlabeled response, we combine its scorer prediction
with the weighted calibration scores to construct a calibrated
reward range with $\alpha=0.1$.

\noindent\textbf{Evaluation.}
We evaluate performance on both in-distribution (ID) mathematical
benchmarks and out-of-distribution (OOD) benchmarks outside the
labeled mathematics domain. The ID benchmarks include AIME 2024, AIME 2025, AMC,
MATH-500 \citep{lightman2024let}, Minerva \citep{lewkowycz2022solving}, and OlympiadBench \citep{he2024olympiadbench}, whereas the OOD
benchmarks include ARC-c \citep{clark2018think}, GPQA-diamond \citep{rein2023gpqa}, and MMLU-Pro.
We report avg@32 on AIME 2024, AIME 2025, and AMC,
and pass@1 on the remaining benchmarks.

\noindent\textbf{Quantitative Analysis.}
Table~\ref{tab:ood_unlabeled} shows that Range-GRPO achieves the
highest ID and OOD averages among
semi-supervised methods.
It also achieves the best accuracy on eight of the nine individual
benchmarks.
Both averages exceed those of our TraPO reproduction
with an entropy coefficient of $0.001$.
Notably, Range-GRPO attains this performance while reserving part of
the same training prompt pool for scorer fitting, calibration, and
density ratio estimation rather than policy optimization.
Despite this auxiliary allocation, its total response budget,
including both policy and judge outputs, is approximately 18.2\%
smaller than that of TraPO.

Detailed prompt allocation and response accounting are provided in
Section~\ref{sec:experimental_details}.
Range-GRPO also achieves higher ID and OOD averages than UARM, which
applies uncertainty weights at the response level.
This result is consistent with the benefit of preserving
interval-dependent pairwise structure when constructing learning
signals.
Range-GRPO also remains competitive with the fully supervised 2K
setting, exceeding its ID average while remaining close on OOD.
Together, these results support the effectiveness of Range-GRPO for
semi-supervised post-training across the evaluated benchmarks.

\begin{wraptable}[16]{r}{0.52\textwidth}
\vspace{-1.2\baselineskip}
\centering

\caption{
\textbf{Uncertainty-aware pairwise weighting can reverse the learning
signal for a correct response.}
On this unlabeled OOD training instance, midpoint scalarization under
Dr.GRPO assigns the correct response R7 a negative advantage because
its midpoint is $0.7384$, below the group mean of $0.7402$.
Range-GRPO ($\lambda$=1) instead yields a positive advantage by weighting pairwise
comparisons according to interval uncertainty.
Correctness labels are shown only for post-hoc interpretation.
}
\label{tab:qualitative-signal-reversal}

\begingroup
\small
\renewcommand{\arraystretch}{1.15}
\setlength{\tabcolsep}{5pt}

\setlength{\aboverulesep}{0pt}
\setlength{\belowrulesep}{0pt}
\setlength{\belowbottomsep}{0pt}

\begin{NiceTabular*}{\linewidth}{
@{\extracolsep{\fill}}
lcc
@{}
}
\CodeBefore
\rectanglecolor{gray!10}{1-1}{1-3}
\Body

\toprule

\textbf{Response}
&
\textbf{Correctness}
&
\textbf{Calibrated range}
\\

\midrule

R1
&
Incorrect
&
$[0.0000,0.1173]$
\\

R2--R6, R8
&
Correct
&
$[0.7082,1.0000]$
\\

\textbf{R7}
&
Correct
&
$[0.4767,1.0000]$
\\

\bottomrule

\end{NiceTabular*}

\endgroup
\vspace{-0.5\baselineskip}
\end{wraptable}

\noindent\textbf{Qualitative Analysis.}
\label{sec:qualitative-interval-scalarization}
To examine the practical benefit of preserving pairwise range information
as analyzed in Section~\ref{sec:pairwise_range_credit}, we compare its
credit assignment with scalar reduction in the Range-GRPO training logs.
Table~\ref{tab:qualitative-signal-reversal} illustrates how
pairwise weighting based on interval widths can reverse the advantage
sign relative to the scalar-reduction baseline on an unlabeled OOD training
instance.
We compare Range-GRPO with midpoint scalarization under Dr.GRPO,
which achieves the second-highest mean test accuracy in Table~\ref{tab:crpotoy-main}.
Both methods use the same eight responses and calibrated intervals.
The midpoint baseline assigns R7 a negative advantage because
its midpoint lies below the group mean.
The six negative comparisons involve wider counterpart intervals
than the positive comparison with R1.
Range-GRPO therefore assigns smaller weights to those six comparisons,
yielding a positive advantage for R7. This example suggests that interval-aware pairwise weighting can produce a learning signal better aligned with response correctness.

\section{Conclusion}
We present Range-GRPO, a semi-supervised post-training framework
that learns from unlabeled prompts with limited labeled supervision.
The proposed objective assigns relative credit to each response
through pairwise comparisons of reward intervals.
Our theoretical analysis shows that Range-GRPO includes Dr.GRPO as a
special case under zero-width intervals, while its pairwise credit
assignment allows range information to affect both the direction and
magnitude of the learning signal. Empirically, Range-GRPO outperforms the other semi-supervised methods
in both ID and OOD average performance and achieves the best result
on eight of the nine individual benchmarks. Together, these findings highlight the value of using interval
relations to determine not only the strength of a learning signal
but also its direction.
Beyond the evaluated tasks, future work could extend this framework
to domains where reward supervision is uncertain and expert
annotations are limited.

\bibliography{main}
\bibliographystyle{iclr2027_conference}

\appendix
\clearpage
\newsavebox{\toyLeftTableBox}
\newsavebox{\toyRightTableBox}
\newlength{\toyRowPad}

\newcommand{\toyPaddedRow}[1]{%
  \noalign{\vskip\toyRowPad}#1\\\noalign{\vskip\toyRowPad}%
}

\newcommand{\toyAllocationTable}{%
  \begin{tabularx}{\linewidth}{
    @{}l>{\raggedright\arraybackslash}Xr@{}
  }
    \toprule
    \toyPaddedRow{Split & Role & Prompts}
    \midrule
    \toyPaddedRow{Labeled & Scorer / source DRE & 100}
    \toyPaddedRow{& Calibration & 100}
    \addlinespace[2pt]
    \toyPaddedRow{Unlabeled & Policy & 900}
    \toyPaddedRow{& Target DRE & 100}
    \midrule
    \toyPaddedRow{Validation & Model selection & 400}
    \toyPaddedRow{Test & Evaluation & 1,000}
    \bottomrule
  \end{tabularx}%
}

\newcommand{\toyHyperparameterTable}{%
  \begin{tabularx}{\linewidth}{
    @{}>{\raggedright\arraybackslash}Xrr@{}
  }
    \toprule
    \toyPaddedRow{Method & LR & KL}
    \midrule
    \toyPaddedRow{LB (GRPO) & $4\times10^{-5}$ & $0.1$}
    \toyPaddedRow{Midpoint (GRPO) & $5\times10^{-5}$ & $0$}
    \toyPaddedRow{UB (GRPO) & $5\times10^{-5}$ & $0.1$}
    \toyPaddedRow{Scalar Range (Dr.GRPO) & $5\times10^{-5}$ & $0$}
    \toyPaddedRow{UARM & $5\times10^{-5}$ & $0.01$}
    \toyPaddedRow{
      \textbf{Range-GRPO} ($\lambda=0.2$)
      & $5\times10^{-5}$ & $0$
    }
    \bottomrule
  \end{tabularx}%
}

\section{Notation}
\label{sec:notation}

\setlength{\tabcolsep}{6pt}
\renewcommand{\arraystretch}{1.15}
\small

\begin{longtable}{@{}p{0.24\textwidth}p{0.72\textwidth}@{}}
\caption{\textbf{Notation used throughout the paper.}
Symbols are grouped by their roles and ordered approximately by their first appearance in the paper.}
\label{tab:notation}\\

\toprule
\textbf{Notation} & \textbf{Meaning} \\
\midrule
\endfirsthead

\multicolumn{2}{@{}l}{\textbf{Table \thetable\ (continued).} Notation used throughout the paper.} \\
\toprule
\textbf{Notation} & \textbf{Meaning} \\
\midrule
\endhead

\midrule
\multicolumn{2}{r}{Continued on next page} \\
\endfoot

\bottomrule
\endlastfoot

\multicolumn{2}{@{}l}{\textit{GRPO and policy optimization}} \\
\midrule
$q,\;P_Q$
& Prompt and prompt distribution. \\

$G,\;G',\;o_i,\;o_{i,k}$
& Policy and target DRE responses per prompt, response $i$,
and token $k$ of response $i$. \\

$\pi_\theta,\;\pi_{\theta_{\mathrm{old}}},\;\pi_{\mathrm{ref}},\;\pi_t$
& Parameterized policy, old rollout policy, fixed reference policy, and current policy $\pi_t:=\pi_{\theta_t}$. \\

$r_i,\;\bar r,\;\sigma_r$
& Response reward, group-mean reward, and group reward standard deviation. \\

$A_i,\;A_i^{\mathrm{Dr.GRPO}}$
& GRPO and Dr.GRPO response-level advantages. \\

$\rho_{i,k},\;D_{i,k}^{\mathrm{KL}}$
& PPO likelihood ratio and token-level KL penalty. \\

$\epsilon,\;\beta$
& PPO clipping parameter and KL regularization coefficient. \\

\midrule
\multicolumn{2}{@{}l}{\textit{Reward-range construction}} \\
\midrule

$t,\;X=(q,o),\;Y$
& Optimization iteration, prompt-response pair, and reference reward. \\

$P_{\mathrm{src}},\;P_t$
& Source and iteration-$t$ target distributions over $(X,Y)$. \\

$Z=\phi(X),\;\mathcal Z$
& Low-cardinality judgment state, state-extraction mapping, and state space. \\

$P_{\mathrm{src}}^Z,\;P_t^Z$
& Source and target marginal distributions of the judgment state. \\

$s_Z,\;E_j^Z$
& Fixed state scorer and absolute calibration nonconformity score. \\

$C,\;\widehat p_t^{\mathrm{tgt}}(z)$
& Domain indicator ($0$: source; $1$: target) and estimated
posterior $\Pr(C=1\mid Z=z)$. \\

$\widehat w_t(z),\;m_{\mathrm{src}},\;m_t$
& Target-to-source density-ratio weight and source/target sample counts used for its estimation. \\

$\alpha,\;\widehat\tau_{\alpha,t}(z)$
& Miscoverage level and calibrated reward-space radius. \\

$\widehat{\mathcal I}_{\alpha,t}(X)$
& Calibrated reward interval for prompt-response pair $X$. \\

\midrule
\multicolumn{2}{@{}l}{\textit{Range-GRPO credit assignment}} \\
\midrule

$\mathcal I_i=[L_i,U_i]$
& Reward interval for response $i$, with lower and upper endpoints $L_i$ and $U_i$. \\

$c_i,\;h_i$
& Interval center $c_i=(L_i+U_i)/2$ and half-width $h_i=(U_i-L_i)/2$. \\

$\lambda,\;\omega_{ij}$
& Reference-width parameter and pairwise reliability weight $\omega_{ij}=\lambda/(h_i+h_j+\lambda)$. \\

$g_{ij}$
& Pairwise credit signal $g_{ij}=\omega_{ij}(c_i-c_j)$. \\

$A_i^{\mathrm{RANGE}}$
& Range-GRPO response-level advantage obtained by aggregating pairwise signals. \\

$\xi,\;r_i^{\mathrm{pt}}$
& Relative position for scalar interval reduction and the resulting point reward. \\

$\mathcal G,\;\bar r^{\mathrm{pt}}(\mathcal G),\;A_i^{\mathrm{pt}}(\mathcal G)$
& Rollout group, its mean point reward, and point-reduction advantage. \\

$A_i^{\mathrm{wt}}(\mathcal G),\;\eta_i(\mathcal G)$
& Generic positively reweighted point advantage and its scaling factor. \\

$\eta_i^{\mathrm{UARM}},\;\sigma_{\mathrm{signal}},\;\sigma_{\mathrm{noise},i}$
& UARM scaling factor and its signal/noise quantities. \\

\midrule
\multicolumn{2}{@{}l}{\textit{Training procedure}} \\
\midrule

$\mathcal D_U^{\mathrm{pol}},\;\mathcal D_L^{\mathrm{src}},\;\mathcal D_L^{\mathrm{cal}},\;\mathcal D_U^{\mathrm{dre}}$
& Unlabeled policy, labeled source, labeled calibration, and unlabeled target-DRE prompt sets. \\

$J,\;\mathcal Y^{\mathrm{src}},\;\mathcal Y^{\mathrm{cal}}$
& Frozen judge and source/calibration reward collections. \\

$\mathcal Z^{\mathrm{src}},\;\mathcal Z^{\mathrm{cal}},\;n_{\mathrm{cal}}$
& Source and calibration state collections and number of calibration rollouts. \\

$\mathcal Q_t,\;\mathcal Z_t^{\mathrm{pol}},\;\mathcal Z_t^{\mathrm{dre}},\;T$
& Policy-prompt batch, policy/DRE state collections at iteration $t$, and total number of training iterations. \\

\midrule
\multicolumn{2}{@{}l}{\textit{Appendix-specific notation}} \\
\midrule

$Z_i=(z_{i1},\ldots,z_{i5})$
& Five-dimensional judgment state used in the main experiments. \\

$R_j,\;\psi(v),\;\widehat q_{\alpha,t}(Z)$
& Normalized nonconformity score, residual-scale function, and weighted cutoff for normalized conformal calibration. \\

$\zeta$
& Signal-variance stabilizer used in the UARM baseline. \\

\end{longtable}

\normalsize

\section{Toy Experiment}
\label{sec:crpotoy-appendix}

\subsection{Data and Training}
\label{sec:crpotoy-data}
\label{sec:crpotoy-optimization}

\noindent\textbf{Task.}
Following SSPO \citep{lee2026semi}, we construct a synthetic dataset
using words from the Natural Language Toolkit (NLTK) word list \citep{loper2002nltk}. We define a task-specific ranking rule for the candidate words.
Each prompt contains ten candidate words labeled A--J, and the policy
selects one letter.
We first sort the words by increasing length, preserving the input
order for words of equal length.
We then reorder only the three shortest words alphabetically,
treating uppercase and lowercase letters alike.
The remaining seven words retain their order.
A selection is correct if $\operatorname{rank}(o)=1$, while the reward
is determined by
$r(o)=(11-\operatorname{rank}(o))/10$.
In the example shown in Table~\ref{tab:crpotoy-example},
the final order begins with \emph{nominee}, \emph{savoy},
and \emph{wanton}.
Training, validation, and test sets have disjoint prompt compositions.

\noindent\textbf{Data allocation.}
For policy training and auxiliary range construction, we use
1,200 synthetic prompts: 200 labeled and 1,000 unlabeled.
We allocate these prompts to scorer fitting, calibration,
density ratio estimation (DRE), and policy training.
Table~\ref{tab:crpotoy-roles} summarizes this allocation together
with the separate validation and test sets.
All methods use the same prompt partitions.

\begin{table}[!ht]
\centering
\renewcommand{\arraystretch}{1.11}
\setlength{\tabcolsep}{2.5pt}

\toyPreparePair{10}{7}{%
  \begin{tabular*}{\linewidth}{@{\extracolsep{\fill}}llrrr@{}}
    \toprule
    Choice & Word & Length & Rank & Reward \\
    \midrule
    \toyBodyRow{A & intoxicating & 12 & 8 & 0.3}
    \toyBodyRow{B & transcendental & 14 & 10 & 0.1}
    \toyBodyRow{C & tenacious & 9 & 5 & 0.6}
    \toyBodyRow{D & nocturne & 8 & 4 & 0.7}
    \toyBodyRow{E & knickerbocker & 13 & 9 & 0.2}
    \toyBodyRow{F & savoy & 5 & 2 & 0.9}
    \toyBodyRow{G & hypodermic & 10 & 6 & 0.5}
    \toyBodyRow{H & blossomtime & 11 & 7 & 0.4}
    \toyBodyRow{I & nominee & 7 & 1 & 1.0}
    \toyBodyRow{J & wanton & 6 & 3 & 0.8}
    \bottomrule
  \end{tabular*}%
}{%
  \begin{tabularx}{\linewidth}{
    @{}l>{\raggedright\arraybackslash}Xr@{}
  }
    \toprule
    Split & Role & Prompts \\
    \midrule
    \toyBodyRow{Labeled & Scorer / source DRE & 100}
    \toyBodyRow{& Calibration & 100}
    \midrule
    \toyBodyRow{Unlabeled & Policy & 900}
    \toyBodyRow{& Target DRE & 100}
    \midrule
    \toyBodyRow{Validation & Model selection & 400}
    \midrule
    \toyBodyRow{Test & Evaluation & 1,000}
    \midrule
    \toyBodyRow{
      \multicolumn{2}{@{}l}{\textbf{Total}}
      & \textbf{2,600}
    }
    \bottomrule
  \end{tabularx}%
}

\toyCaptionPair
  {Example candidates with their ranks and hidden rewards.}
  {tab:crpotoy-example}
  {Prompt allocation, separating policy learning from range construction.}
  {tab:crpotoy-roles}

\end{table}

\noindent\textbf{Training and model selection.}
All methods train for ten epochs without warmup or labeled policy updates.
Table~\ref{tab:crpotoy-optimizer} summarizes the training settings.
We select learning rates from $\{1,2,4,5\}\times10^{-5}$ and KL
coefficients from $\{0,0.01,0.1\}$.
For Range-GRPO, we select $\lambda=0.2$ from the grid
$\{0.2,0.4,0.6,0.8,1\}$.
We select the best checkpoint in each run based on validation accuracy, then choose hyperparameters using the mean of these best validation accuracies across three seeds.

\begin{table}[!ht]
\centering
\caption{
\textbf{Policy training configuration and selected hyperparameters.}
(a) summarizes the shared training configuration for all methods.
(b) lists the selected learning rate and KL coefficient for each method.
We select the hyperparameters by mean best validation accuracy across three seeds.
LR denotes the learning rate.
}
\label{tab:crpotoy-optimizer}
\label{tab:crpotoy-hp}

\small
\renewcommand{\arraystretch}{1.08}
\setlength{\tabcolsep}{3pt}

\toyPreparePair{10}{8}{%
  \begin{tabularx}{\linewidth}{
    @{}l>{\raggedright\arraybackslash}X@{}
  }
    \toprule
    Setting & Value \\
    \midrule
    \toyBodyRow{Model & GPT-2-small}
    \toyBodyRow{Batch size & 16}
    \toyBodyRow{Responses per prompt & 8}
    \toyBodyRow{Epochs & 10}
    \toyBodyRow{Warmup epochs & 0}
    \toyBodyRow{Optimizer & AdamW}
    \toyBodyRow{Clipping & 0.2}
    \toyBodyRow{Entropy coefficient & 0.01}
    \toyBodyRow{Checkpoint & Best validation accuracy}
    \toyBodyRow{Seeds & 6200, 6201, 6202}
    \bottomrule
  \end{tabularx}%
}{%
  \begin{tabularx}{\linewidth}{
    @{}>{\raggedright\arraybackslash}Xrr@{}
  }
    \toprule
    Method & LR & KL \\
    \midrule
    \toyBodyRow{
      LB (GRPO) & $4\times10^{-5}$ & $0.1$
    }
    \toyBodyRow{
      Midpoint (GRPO) & $5\times10^{-5}$ & $0$
    }
    \toyBodyRow{
      UB (GRPO) & $5\times10^{-5}$ & $0.1$
    }
    \toyBodyRow{
      LB (Dr.GRPO) & $5\times10^{-5}$ & $0$
    }
    \toyBodyRow{
      Midpoint (Dr.GRPO) & $5\times10^{-5}$ & $0$
    }
    \toyBodyRow{
      UB (Dr.GRPO) & $5\times10^{-5}$ & $0$
    }
    \toyBodyRow{
      UARM & $5\times10^{-5}$ & $0.01$
    }
    \toyBodyRow{
      \textbf{Range-GRPO} ($\lambda=0.2$)
      & $5\times10^{-5}$ & $0$
    }
    \bottomrule
  \end{tabularx}%
}

\toyPanelPair
  {(a) Training configuration}
  {(b) Selected hyperparameters}

\end{table}

\subsection{Learning Signals}
\label{sec:crpotoy-uarm-method}
\label{sec:crpotoy-dynamics}
\label{sec:crpo-no-warmup-appendix}

\noindent\textbf{Advantage construction.}
All methods in this toy comparison use the interval construction
procedure in Section~\ref{sec:crpotoy-conformal} but apply different
rules to compute advantages from the resulting intervals.
The scalar baselines use the lower bound (LB), upper bound (UB),
or midpoint of each interval as the reward and compute either
GRPO or Dr.GRPO advantages.
Range-GRPO uses the objective defined in
Section~\ref{sec:range_grpo_objective} with $\lambda=0.2$
and no additional advantage normalization.

We adapt UARM \citep{pan2026uncertainty} by using the midpoint
of each interval as the reward estimate for the corresponding response.
The shared interval construction procedure replaces the original quantile
reward model so that all advantage rules use the same source of uncertainty.
We retain the published advantage formula and use a signal variance
stabilizer of $10^{-6}$.

\begin{figure}[!ht]
\centering
\includegraphics[width=.80\linewidth]{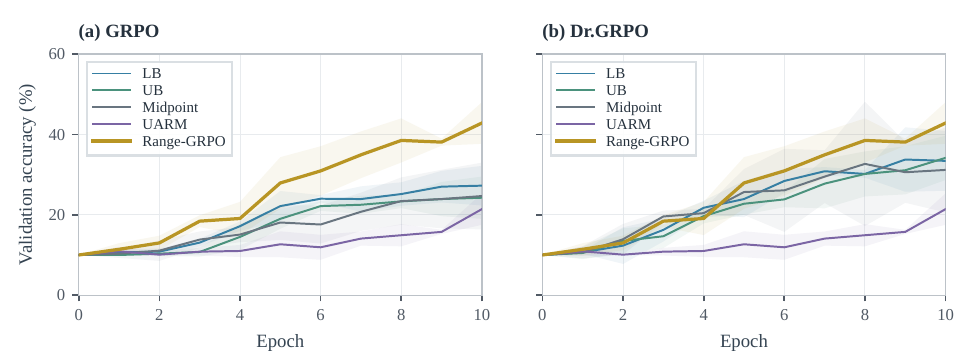}
\caption{
\textbf{Validation accuracy at the selected hyperparameter settings.}
Curves show the mean $\pm$ sample standard deviation across three seeds.
We show scalar baselines with GRPO and Dr.GRPO advantages separately
and include the same UARM and Range-GRPO curves in both comparisons.
}
\label{fig:crpo-no-warmup-validation}
\label{fig:crpotoy-curves}
\end{figure}

\noindent\textbf{Training dynamics.}
Figure~\ref{fig:crpo-no-warmup-validation} shows the validation curves
for each method at the selected hyperparameter settings.
Range-GRPO shows a stronger upward trend in validation accuracy,
particularly relative to the scalar baselines using GRPO advantages.

\subsection{Range Construction}
\label{sec:crpotoy-conformal}
\label{sec:crpotoy-budget}

\noindent\textbf{Response features.}
To simplify state construction in the toy task, we compute $Z$
directly from word features rather than using an LLM judge.
For each selected word, $Z$ contains five fixed features:
Brown corpus frequency, Brown genre count, WordNet synset count,
WordNet noun status, and surprisal of the first token under GPT-2
after a fixed end-of-sequence context.
We apply $\log(1+x)$ to the first three features and use fixed
standardization for all five.

\noindent\textbf{Scorer and calibration scores.}
We adapt the RoWCI scorer \citep{noh2026reliable} to the toy task
using an XGBoost regressor that predicts rewards from $Z$.
Using responses from the initial pretrained policy, we fit the scorer
and compute calibration nonconformity scores on their respective prompt
partitions.
The fitted scorer and calibration scores remain fixed for the
rest of training.
All methods share the fitted scorer, calibration scores,
and DRE source data.

\noindent\textbf{Updates during training.}
At each update using unlabeled responses, a random forest estimates
density ratios from a fixed source feature set and features extracted
from new target responses sampled by each method's current policy
on the shared target prompt pool.
These ratios reweight the fixed calibration scores to account for
changes in the response distribution without new target reward labels.
For numerical stability, we cap the estimated density-ratio weights
at their 95th percentile.
The weighted cutoff also accounts for the weight of the query response.
We clip intervals to $[0,1]$ and use this full range when the cutoff
is infinite.
We exclude unlabeled rewards from policy optimization and use them
for coverage diagnostics at a nominal level of $90\%$.

\noindent\textbf{Response budget.}
Table~\ref{tab:crpotoy-budget}(b) details the response budget for
policy training, scorer fitting, calibration, and target DRE,
excluding validation and test evaluations.
Source DRE reuses the scorer responses and therefore requires no
additional generations.
Because we compute $Z$ directly from word features, the toy experiment
does not require an LLM judge.

\begin{table}[!ht]
\centering
\caption{
\textbf{Range construction settings and response budget.}
(a) summarizes the range construction settings.
(b) reports the response budget for policy training and range construction.
}
\label{tab:crpotoy-range}
\label{tab:crpotoy-budget}

\small
\renewcommand{\arraystretch}{1.08}
\setlength{\tabcolsep}{3pt}

\toyPreparePair{8}{5}{%
  \begin{tabularx}{\linewidth}{
    @{}>{\raggedright\arraybackslash}X
    >{\raggedleft\arraybackslash}X@{}
  }
    \toprule
    Setting & Value \\
    \midrule
    \toyBodyRow{Scorer & XGBoost regressor}
    \toyBodyRow{Trees / depth & 200 / 4}
    \toyBodyRow{DRE & Random forest}
    \toyBodyRow{Trees / depth & 200 / 8}
    \toyBodyRow{DRE target per update & 255 responses}
    \toyBodyRow{Refresh & Each unlabeled update}
    \toyBodyRow{Nominal coverage & 90\%}
    \toyBodyRow{Weight cap & 95th percentile}
    \bottomrule
  \end{tabularx}%
}{%
  \begin{tabularx}{\linewidth}{
    @{}>{\raggedright\arraybackslash}Xr@{}
  }
    \toprule
    Category & Responses \\
    \midrule
    \toyBodyRow{Unlabeled policy & 72,000}
    \toyBodyRow{Scorer & 5,600}
    \toyBodyRow{Calibration & 5,600}
    \toyBodyRow{Target DRE & 145,350}
    \midrule
    \toyBodyRow{
      \textbf{Total} & \textbf{228,550}
    }
    \bottomrule
  \end{tabularx}%
}

\toyPanelPair
  {(a) Range construction settings}
  {(b) Response budget}

\end{table}

\FloatBarrier

\section{Experimental Details}
\label{sec:experimental_details}

We adapt the experimental setting of Table~2 in
TraPO~\citep{yang2026trapo} for both Range-GRPO and
UARM~\citep{pan2026uncertainty}.
For all methods, we initialize the policy from
Qwen2.5-Math-7B~\citep{yang2024qwen2}. Following TraPO, we select the checkpoint with the highest mean validation accuracy across AIME, AMC, MATH, Minerva, and OlympiadBench. Results under our experimental setting are reported from a single training run per configuration.

\subsection{TraPO}

We reproduce TraPO using its official implementation.
The entropy coefficient differs between the paper and the official
implementation: the paper uses $0.01$, whereas the released configuration
uses $0.001$. We therefore evaluate both entropy coefficients while keeping all other settings fixed.
As shown in Table~\ref{tab:trapo_entropy}, an entropy coefficient of $0.001$
outperforms $0.01$ on all benchmarks except GPQA, improving the average ID
and OOD performance by $2.6$ and $1.2$ percentage points, respectively.
We therefore report the results with an entropy coefficient of $0.001$
in the main table.

For budget accounting, we follow the TraPO training procedure,
which uses a warmup of 10 epochs followed by 5 epochs that combine
labeled and unlabeled signals.
TraPO generates rollouts for both labeled and unlabeled prompts in every
training epoch.
During warmup, only labeled rollouts contribute to policy updates,
while unlabeled rollouts accumulate pass rate trajectories.
In the subsequent stage, TraPO uses these trajectories to select reliable
unlabeled samples for policy updates.
We therefore count both labeled and unlabeled rollouts across all
15 training epochs.
Table~\ref{tab:baseline-response-budget} summarizes the resulting
response budget.

\begin{table}[t]
\centering
\caption{
\textbf{Performance of reproduced TraPO configurations with different entropy coefficients.}
All training methods use 1,024 (1K) ID and 1,024 (1K) OOD prompts.
GPQA$^{\dagger}$ denotes GPQA-diamond.
Bold and underlined scores indicate the best and second-best results, respectively.
}
\label{tab:trapo_entropy}

\fontsize{8.5}{11}\selectfont\normalfont
\setlength{\tabcolsep}{1.0pt}
\renewcommand{\arraystretch}{1.30}

\setlength{\aboverulesep}{0pt}
\setlength{\belowrulesep}{0pt}
\setlength{\belowbottomsep}{0pt}

\resizebox{\linewidth}{!}{%
\begin{tabular}{
    @{}
    l
    c
    c
    c
    c
    c
    @{\hspace{4pt}}
    c
    @{\hspace{7pt}}
    c
    c
    c
    @{\hspace{4pt}}
    c
    @{}
}

\toprule
& \multicolumn{6}{c}{\textbf{In-Distribution}}
& \multicolumn{4}{c}{\textbf{Out-of-Distribution}} \\
\cmidrule(lr){2-7}
\cmidrule(lr){8-11}

Model / Method
& AIME 24/25
& AMC
& MATH-500
& Minerva
& Olympiad
& \textbf{Avg.}
& ARC-c
& GPQA$^{\dagger}$
& MMLU-Pro
& \textbf{Avg.} \\

\midrule

TraPO (entropy=0.01)
& \underline{17.3}/\underline{10.3}
& \underline{52.6}
& \underline{77.6}
& \underline{33.8}
& \underline{36.3}
& \underline{38.0}
& \underline{78.8}
& \textbf{36.9}
& \underline{39.5}
& \underline{51.7} \\

TraPO (entropy=0.001)
& \textbf{17.7}/\textbf{12.2}
& \textbf{55.9}
& \textbf{79.2}
& \textbf{37.5}
& \textbf{41.3}
& \textbf{40.6}
& \textbf{80.4}
& \underline{34.8}
& \textbf{43.5}
& \textbf{52.9} \\

\bottomrule
\end{tabular}%
}

\end{table}

\begin{table}[htbp]
\centering
\caption{
\textbf{Response budget for the main experiments.}
The table compares the response budget for TraPO, UARM, and Range-GRPO.
For TraPO, unlabeled rollouts include those generated during warmup.
The UARM budget includes RM processing, whereas the Range-GRPO budget
includes judge generations.
}
\label{tab:baseline-response-budget}

\small
\setlength{\tabcolsep}{3pt}
\renewcommand{\arraystretch}{1.06}

\begin{tabular*}{\linewidth}{
    @{\extracolsep{\fill}}lrrr@{}
}
\toprule
Category & TraPO & UARM & Range-GRPO \\
\midrule
Labeled policy
& 122,880 & 92,160 & 92,160 \\
Unlabeled rollouts
& 122,880 & 40,960 & 37,120 \\
RM training
& 0 & 28,560 & 0 \\
RM calibration
& 0 & 7,280 & 0 \\
RM processing
& 0 & 76,800 & 0 \\
Scorer fitting
& 0 & 0 & 1,280 \\
Calibration
& 0 & 0 & 768 \\
Target DRE
& 0 & 0 & 57,600 \\
Judge generations
& 0 & 0 & 12,096 \\
\midrule
\textbf{Total allocation}
& \textbf{245,760}
& \textbf{245,760}
& \textbf{201,024} \\
\bottomrule
\end{tabular*}

\end{table}

\subsection{UARM}
\label{sec:baseline-implementation}

\paragraph{Reward Model Training.}
We use a frozen Qwen2.5-Math-7B checkpoint~\citep{yang2024qwen2}
as the RM backbone and train only a quantile head. The quantile head jointly predicts multiple reward quantiles
for each prompt-response pair.
For each prompt reserved for RM training, validation, or calibration,
the initial policy generates 140 responses.
Correctness checks against the ground truth provide binary reward
targets: one for a correct answer and zero otherwise.
Within each configuration, we retain the RM checkpoint with
the lowest validation loss.
Among configurations, we select the one with the highest mean
validation AUC of the predicted median rewards.
We use a separate labeled subset for calibration.
Both the RM and calibration parameters remain fixed during
policy training.

\paragraph{Policy Training.}
The experiment uses 2,048 prompts, comprising 1,024 labeled
and 1,024 unlabeled prompts.
Table~\ref{tab:main-data-allocation} shows the allocation
of labeled prompts to disjoint subsets.
Following the warmup setting for other baselines in Table~\ref{tab:ood_unlabeled},
training begins with ten epochs of labeled updates.
The next five epochs combine labeled and unlabeled updates,
giving 15 epochs in total.
For labeled updates, we center the binary correctness rewards
by their group mean, as in Dr.GRPO~\citep{liu2025understanding}.
Unlabeled updates use UARM advantages within the GRPO objective.
We set the signal variance stabilizer to $\zeta=10^{-6}$.
For policy checkpoint selection, we choose the checkpoint with
the highest mean accuracy across the five validation benchmarks.
Table~\ref{tab:baseline-uarm-settings} lists the main training settings.

\begin{table}[htbp]
\centering
\caption{
\textbf{UARM training configuration for the cross-domain setting.}
The table reports the policy training settings in (a) and the reward model
training and calibration settings in (b).
}
\label{tab:baseline-uarm-settings}

\begin{minipage}[t]{.48\linewidth}
\vspace{0pt}
\centering
\small
\textbf{(a) Policy training}\par\smallskip
\setlength{\tabcolsep}{3pt}
\renewcommand{\arraystretch}{1.20}

\begin{tabularx}{\linewidth}{
    @{}>{\raggedright\arraybackslash}Xr@{}
}
\toprule
Setting & Value \\
\midrule
Model & Qwen2.5-Math-7B \\
Learning rate & $10^{-6}$ \\
Batch size & 64 prompts \\
Responses per prompt & 8 \\
Epochs & 15 \\
Warmup epochs & 10 \\
\bottomrule
\end{tabularx}
\end{minipage}\hfill%
\begin{minipage}[t]{.48\linewidth}
\vspace{0pt}
\centering
\small
\textbf{(b) RM training and calibration}\par\smallskip
\setlength{\tabcolsep}{3pt}
\renewcommand{\arraystretch}{1.06}

\begin{tabularx}{\linewidth}{
    @{}>{\raggedright\arraybackslash}Xr@{}
}
\toprule
Setting & Value \\
\midrule
Learning rate & $10^{-4}$ \\
Batch size & 1,024 responses \\
Quantile outputs & 11 \\
Hidden dimensions & 256, 64 \\
Maximum epochs & 600 \\
Patience & 30 \\
Miscoverage $\alpha$ & 0.1 \\
\bottomrule
\end{tabularx}
\end{minipage}
\end{table}

\paragraph{Response Budget.}
For budget accounting, we count each generated response and each
response processed by the RM as one budget unit.
RM processing includes offline feature extraction and
unlabeled response scoring.
This accounting convention is used only for budget allocation.
It does not assume that response generation and RM processing
require the same computational cost.

\begin{table}[htbp]
\centering
\caption{
\textbf{Data allocation for UARM and Range-GRPO in the main experiments.}
(a) summarizes the UARM allocation across policy learning and
RM training, validation, and calibration.
(b) details the Range-GRPO allocation across policy learning,
scorer fitting, calibration, source DRE, and target DRE.
}
\label{tab:main-data-allocation}
\label{tab:range-grpo-data-roles}

\small
\setlength{\tabcolsep}{3pt}
\renewcommand{\arraystretch}{1.08}

\begin{minipage}[t]{.48\linewidth}
\vspace{0pt}
\centering
\textbf{(a) UARM}\par\smallskip

\begin{tabularx}{\linewidth}{
    @{}>{\raggedright\arraybackslash}Xr@{}
}
\toprule
Role & Prompts \\
\midrule
Labeled policy & 768 \\
RM training & 184 \\
RM validation & 20 \\
Calibration & 52 \\
\midrule
Unlabeled policy & 1,024 \\
\midrule
\textbf{Total} & \textbf{2,048} \\
\bottomrule
\end{tabularx}
\end{minipage}\hfill%
\begin{minipage}[t]{.48\linewidth}
\vspace{0pt}
\centering
\textbf{(b) Range-GRPO}\par\smallskip

\begin{tabularx}{\linewidth}{
    @{}>{\raggedright\arraybackslash}Xr@{}
}
\toprule
Role & Prompts \\
\midrule
Labeled policy & 768 \\
Scorer / source DRE & 160 \\
Calibration & 96 \\
\midrule
Unlabeled policy & 928 \\
Target DRE & 96 \\
\midrule
\textbf{Total} & \textbf{2,048} \\
\bottomrule
\end{tabularx}
\end{minipage}

\end{table}

\subsection{Range-GRPO}
\label{sec:range-grpo-implementation}

\paragraph{Reward range construction.}
We use Qwen3-4B-Instruct-2507
\footnote{\url{https://huggingface.co/Qwen/Qwen3-4B-Instruct-2507}}
as a frozen judge.
The judge extracts a low-cardinality judgment state $Z$ using
general criteria that are not tied to a specific task domain.
Each judge input contains one problem and a group of eight responses.
For each response, we provide its reasoning, submitted answer,
and agreement with other responses.
Reference answers and correctness labels are not provided.
For response $i$, the judge extracts
$Z_i=(z_{i1},\ldots,z_{i5})\in\{0,0.5,1\}^{5}$
according to Table~\ref{tab:range-grpo-z-states}.
The judge produces a single deterministic structured output for each
group of eight responses, with a maximum length of 768 tokens.

\begin{table}[!ht]
\centering
\caption{
\textbf{Five components of the judgment state $Z$.}
The first three states assess the response itself, while the last two
capture agreement with other responses in the same rollout group.
Each state takes a score of 0, 0.5, or 1.
}
\label{tab:range-grpo-z-states}

\footnotesize
\setlength{\tabcolsep}{3pt}
\renewcommand{\arraystretch}{1.18}

\begin{tabularx}{\linewidth}{
    @{}
    >{\centering\arraybackslash}p{.05\linewidth}
    >{\raggedright\arraybackslash}p{.16\linewidth}
    >{\raggedright\arraybackslash}X
    >{\raggedright\arraybackslash}X
    >{\raggedright\arraybackslash}X
    @{}
}
\toprule
State & Criterion & Score 0 & Score 0.5 & Score 1 \\
\midrule

\multicolumn{5}{@{}l}{\textit{Response-level assessment}} \\
\midrule

$z_1$
& Answer verification
& The judge identifies a contradiction when independently checking
the submitted answer against the problem.
& The judge cannot verify or refute the submitted answer
with sufficient confidence.
& The judge independently verifies the submitted answer
against the problem. \\

\midrule

$z_2$
& Reasoning consistency
& The visible reasoning contains a decisive error or conflicts
with the submitted answer.
& The visible reasoning is incomplete or insufficient to establish
the submitted answer.
& The visible reasoning consistently supports the submitted answer. \\

\midrule

$z_3$
& Requirement compliance
& The response omits or violates an explicit task requirement.
& It remains unclear whether all explicit requirements are satisfied.
& The response satisfies all explicit task requirements. \\

\midrule
\midrule

\multicolumn{5}{@{}l}{\textit{Group-level agreement}} \\
\midrule

$z_4$
& Absolute agreement
& No other response gives the same answer.
& One to three other responses give the same answer.
& Four to seven other responses give the same answer. \\

\midrule

$z_5$
& Relative agreement
& Fewer responses support this answer than the most supported alternative.
& This answer and the most supported alternative receive equal support.
& More responses support this answer than any alternative. \\

\bottomrule
\end{tabularx}
\end{table}

For the main experiments, we follow the range construction procedure
in Algorithm~\ref{alg:range-grpo} with two practical implementation choices.
First, to align with the TraPO warmup schedule, we treat the policy
immediately after labeled warmup as $\pi_0$.
At this stage, $\pi_0$ generates the source and calibration responses
for scorer fitting and calibration.
We keep the fitted scorer and calibration nonconformity scores fixed
throughout subsequent training.

Second, we normalize the calibration nonconformity scores so that
differences in scorer predictions are reflected more clearly during
range construction.
For a scorer prediction $s_Z(Z)\in[0,1]$, we define
\begin{equation}
R_j
=
\frac{|Y_j-s_Z(Z_j)|}
     {\psi\!\left(s_Z(Z_j)\right)},
\qquad
\psi(v)=\max\{2\min(v,1-v),0.05\}.
\label{eq:range-grpo-exp-residual-scale}
\end{equation}
For the same positive prediction error, the normalized
nonconformity score increases as the scorer prediction approaches
$0$ or $1$, until $\psi(v)$ reaches its floor of $0.05$.
This makes the nonconformity score sensitive to the scorer prediction
in addition to the absolute prediction error.

We compute the weighted cutoff $\widehat q_{\alpha,t}(Z)$ from the normalized
nonconformity scores and rescale it for each query response as
$\widehat\tau_{\alpha,t}(Z)
= \psi(s_Z(Z))\,\widehat q_{\alpha,t}(Z)$.
Here, $\psi(s_Z(Z))$ sets the query-specific error scale, while
$\widehat q_{\alpha,t}(Z)$ provides the corresponding normalized
conformal cutoff.
We then use $\widehat\tau_{\alpha,t}(Z)$ in the range construction
of Algorithm~\ref{alg:range-grpo}.
Together, normalization and query-specific rescaling allow the
constructed ranges to reflect differences in scorer confidence more
directly. At each unlabeled update, we refit the DRE classifier using the fixed
source states and fresh target states generated by the current policy.
Table~\ref{tab:range-grpo-range-settings} summarizes the main range
construction settings.

\begin{table}[!ht]
\centering
\caption{
\textbf{Range construction settings for the main experiments.}
The scorer and calibration nonconformity scores remain fixed after warmup,
whereas DRE is refreshed at each unlabeled update.
}
\label{tab:range-grpo-range-settings}

\small
\renewcommand{\arraystretch}{1.08}
\setlength{\tabcolsep}{3pt}

\begin{minipage}{.56\linewidth}
\begin{tabularx}{\linewidth}{
    @{}>{\raggedright\arraybackslash}X
    >{\raggedleft\arraybackslash}X@{}
}
\toprule
Setting & Value \\
\midrule
Scorer & XGBoost regressor \\
Trees / depth & 200 / 2 \\
DRE & Random forest \\
Trees / depth & 200 / 4 \\
DRE target per update & 768 responses \\
Refresh & Each unlabeled update \\
Nominal coverage & 90\% \\
Weight cap & 95th percentile \\
\bottomrule
\end{tabularx}
\end{minipage}

\end{table}

\paragraph{Policy Training.}
Table~\ref{tab:main-data-allocation}(b) summarizes the prompt allocation
used for Range-GRPO.
The five subsets remain fixed and disjoint at the prompt level;
scorer fitting and source DRE share the same responses, with correctness
labels used only for scorer fitting.

We train for ten labeled warmup epochs followed by five active epochs
using both labeled and unlabeled prompts.
During active training, labeled updates use binary correctness rewards,
while unlabeled updates use Range-GRPO advantages following
Algorithm~\ref{alg:range-grpo}.
Table~\ref{tab:range-grpo-settings} summarizes the main training settings.

\begin{table}[!ht]
\centering
\caption{
\textbf{Range-GRPO policy training settings for the main experiments.}
(a) summarizes the policy training configuration.
(b) lists the main optimization and Range-GRPO settings.
LR denotes the learning rate.
}
\label{tab:range-grpo-settings}

\small
\setlength{\tabcolsep}{3pt}

\begin{minipage}[t]{.48\linewidth}
\vspace{0pt}
\centering
\textbf{(a) Policy training}\par\smallskip

\renewcommand{\arraystretch}{1.25}
\begin{tabularx}{\linewidth}{
    @{}>{\raggedright\arraybackslash}Xr@{}
}
\toprule
Setting & Value \\
\midrule
Batch size & 64 prompts \\
Responses per prompt & 8 \\
Warmup epochs & 10 \\
Active epochs & 5 \\
Optimizer & AdamW \\
PPO clipping & 0.2 \\
\bottomrule
\end{tabularx}
\end{minipage}\hfill%
\begin{minipage}[t]{.48\linewidth}
\vspace{0pt}
\centering
\textbf{(b) Main configuration}\par\smallskip

\renewcommand{\arraystretch}{1.25}
\begin{tabularx}{\linewidth}{
    @{}>{\raggedright\arraybackslash}Xr@{}
}
\toprule
Setting & Value \\
\midrule
LR & $3\times10^{-6}$ \\
Entropy coefficient & 0.01 \\
KL coefficient & 0 \\
$\lambda$ & 1 \\
Maximum prompt tokens & 3,072 \\
Maximum response tokens & 4,096 \\
\bottomrule
\end{tabularx}
\end{minipage}

\end{table}

\paragraph{Response Budget.}
Table~\ref{tab:baseline-response-budget} details the response budget
for policy training and reward range construction.
We include both policy rollouts and judge generations when computing
the Range-GRPO response budget.
The resulting budget is 201,024 units, which is 18.2\% lower than
the 245,760 units used for TraPO and UARM.
Under this reduced budget, Range-GRPO still achieves strong overall
performance, as shown in Table~\ref{tab:ood_unlabeled}.

\paragraph{Wall-clock Runtime.}
Table~\ref{tab:runtime-comparison} reports the observed wall-clock
runtime for the main experiments.
All runs use NVIDIA RTX PRO 6000 Blackwell Server Edition GPUs.
The GPU count refers to the allocation during active training.
UARM uses four GPUs for policy training and one dedicated GPU for
the reward model (RM), whereas Range-GRPO shares four GPUs between
policy and judge phases.

\begin{table}[!ht]
\centering
\caption{
\textbf{Range-GRPO maintains comparable wall-clock training time
despite the additional judge phase.}
Each run consists of 10 warmup epochs followed by 5 active epochs.
Phase durations and total training times are independently rounded
to one decimal place.
}
\label{tab:runtime-comparison}

\begingroup
\small
\renewcommand{\arraystretch}{1.15}
\setlength{\tabcolsep}{5pt}

\begin{tabular*}{\linewidth}{
    @{\extracolsep{\fill}}
    lcrrr
    @{}
}
\toprule
\textbf{Method}
& \textbf{Active GPUs}
& \textbf{Warmup (h)}
& \textbf{Active (h)}
& \textbf{Training (h)} \\
\midrule

TraPO (entropy=0.01)
& 4
& 12.8
& 10.6
& 23.4 \\

TraPO (entropy=0.001)
& 4
& 12.6
& 7.0
& 19.7 \\

UARM
& 5
& 6.5
& 9.9
& 16.4 \\

Range-GRPO (ours)
& 4
& 6.3
& 10.0
& 16.3 \\

\bottomrule
\end{tabular*}
\endgroup
\end{table}

Training time includes initialization, rollout generation, policy
updates, and validation.
Range-GRPO training time also accounts for judge loading and calls,
scorer fitting, calibration, and DRE updates.
UARM requires an additional 2.6\,h for RM preparation before policy
training, resulting in a total wall-clock time of 19.0\,h when this
preparation is included.

TraPO (entropy=0.01) has a longer active phase than
TraPO (entropy=0.001).
This difference is consistent with their average unlabeled response
lengths of 2,244 and 553 tokens, respectively.
Despite the additional judge phase, Range-GRPO shows a training
wall-clock comparable to UARM while using four rather than five
active GPUs.
This practical efficiency is consistent with the reduced response
budget reported in Table~\ref{tab:baseline-response-budget}.

\FloatBarrier
\end{document}